\documentclass{sig-alternate}

\usepackage{graphicx}
\graphicspath{{.}{img/}}
\usepackage[caption=false,font=normalsize,labelfont=sf,textfont=sf]{subfig}
\usepackage{epstopdf}
\RequirePackage{epsfig}
\DeclareGraphicsExtensions{.eps,.png,.jpg,.pdf,.tif,.tiff,.ps}

\usepackage[svgnames]{xcolor}
\RequirePackage{xspace}
\RequirePackage{eso-pic}
\usepackage{ucs}
\usepackage[utf8x]{inputenc}	%for linux
\usepackage{url}
\usepackage{hyperref}

\newcommand{\squishlist}{
 \begin{list}{$\bullet$}
  { \setlength{\itemsep}{0pt}
     \setlength{\parsep}{3pt}
     \setlength{\topsep}{3pt}
     \setlength{\partopsep}{0pt}
     \setlength{\leftmargin}{1.5em}
     \setlength{\labelwidth}{1em}
     \setlength{\labelsep}{0.5em} } }

\newcommand{\squishlisttwo}{
 \begin{list}{$\bullet$}
  { \setlength{\itemsep}{0pt}
    \setlength{\parsep}{0pt}
    \setlength{\topsep}{0pt}
    \setlength{\partopsep}{0pt}
    \setlength{\leftmargin}{1.5em}
    \setlength{\labelwidth}{1.5em}
    \setlength{\labelsep}{0.5em} } }

\newcommand{\squishend}{
  \end{list}  }

\hypersetup
{
pdftitle={CommentWatcher: An Open Source Web-based platform for analyzing discussions on web forums},%
pdfauthor={Marian-Andrei Rizoiu, Adrien Guille and Julien Velcin},%
pdfsubject={WWW Demo paper of the software CommentWatcher}%
}
\begin{document}

%\conferenceinfo{WOODSTOCK}{'97 El Paso, Texas USA}
%\CopyrightYear{2007} % Allows default copyright year (20XX) to be over-ridden - IF NEED BE.
%\crdata{0-12345-67-8/90/01}  % Allows default copyright data (0-89791-88-6/97/05) to be over-ridden - IF NEED BE.
% --- End of Author Metadata ---

\title{CommentWatcher: An Open Source Web-based platform for analyzing discussions on web forums}

%\subtitle{[Extended Abstract]
%\titlenote{A full version of this paper is available as
%\textit{Author's Guide to Preparing ACM SIG Proceedings Using
%\LaTeX$2_\epsilon$\ and BibTeX} at
%\texttt{www.acm.org/eaddress.htm}}}

\numberofauthors{3}
\author{
% 1st. author
\alignauthor
Marian-Andrei Rizoiu\\
       \affaddr{ERIC Lab, University Lyon2}\\
       \affaddr{Lyon, France}\\
%       \email{\href{mailto:Marian-Andrei.Rizoiu@univ-lyon2.fr}{Marian-Andrei.Rizoiu@univ-lyon2.fr}}
	   \email{Marian-Andrei.Rizoiu@univ-lyon2.fr}
% 2nd. author
\alignauthor
Adrien Guille\\
       \affaddr{ERIC Lab, University Lyon2}\\
       \affaddr{Lyon, France}\\
%       \email{\href{mailto:Adrien.Guille@univ-lyon2.fr}{Adrien.Guille@univ-lyon2.fr}}
	   \email{Adrien.Guille@univ-lyon2.fr}
% 3rd. author
\alignauthor
Julien Velcin\\
       \affaddr{ERIC Lab, University Lyon2}\\
       \affaddr{Lyon, France}\\
%       \email{\href{mailto:Julien.Velcin@univ-lyon2.fr}{Julien.Velcin@univ-lyon2.fr}}
	   \email{Julien.Velcin@univ-lyon2.fr}
}

\maketitle

\begin{abstract}
We present \texttt{CommentWatcher}, an open source tool aimed at analyzing discussions on web forums.
Constructed as a web platform, \texttt{CommentWatcher} features automatic mass fetching of user posts from forum on multiple sites, extracting topics, visualizing the topics as an expression cloud and exploring their temporal evolution.
The underlying social network of users is simultaneously constructed using the citation relations between users and visualized as a %enriched
 graph structure. 
Our platform addresses the issues of the diversity and dynamics of structures of webpages hosting the forums by implementing a parser architecture that is independent of the HTML structure of webpages.
This allows easy on-the-fly adding of new %supported
 websites.
Two types of users are targeted: end users who seek to study the discussed topics and their temporal evolution, and researchers in need of establishing a forum benchmark dataset and comparing the performances of analysis tools.
\end{abstract}


%\terms{Experimentation}

 % NOT required for Proceedings

\section{Introduction}

The Web 2.0 has changed the way users discuss with other users.
%: it allows them to reach people from different geographic locations and different backgrounds.
One of the preferred online discussion environments are the web forums. 
Users can react, post their opinions, discuss and debate any kind of subjects.
The forums are usually thematic (\textit{e.g.} Java programming forums) and new users have access to the past discussion (\textit{e.g.} solutions posted by other users to a specific problem).
Therefore the users become full collaborative participants in the information creation process.
The subjects of discussion between readers are very dynamic and the overall sum of reactions gives a snapshot of the general trends that emerge in the user population.
%It is always interesting to detect the positioning of users concerning certain problems.
%Immediate applications include detecting the population sentiment towards community projects, more efficient publicity \textit{etc.}
At the same time, the way users reply one to another suggests an underlying social network structure.
The forum's ``reply-to'' structural relations can be used to add links between users.
Other types of relations can be added, like the name and textual citations~\cite{FOR11}.
Furthermore, based on such social networks constructed from web forums, adapted graph measures can be used to detect user social roles~\cite{ANO12}.
%, a social network is inferred from TV series forums and adapted graph measures are used to detect user social roles.

\subsection{Current limitations}

These forum data are still ill explored, even if they represent an important source of knowledge.
News articles analysis and micro blogging (\textit{e.g.} Twitter) analysis receive a lot of attention from the community.
There are available tools that perform the analysis of news media~\cite{AME12}, but without treating the social network aspect.
Other tools concentrate on analyzing and visualizing the social dynamics~\cite{GUI13} or detect events~\cite{MAR11} based on twitter data.
To the best of our knowledge, there are no publicly available tools that treat forums, while inferring a social network structure.

Another limitation concerns the forum benchmarks.
There are a multitude of general purpose information retrieval data-sets (\textit{e.g.} the \texttt{ClueWeb12 dataset}\footnote{\url{http://lemurproject.org/clueweb12/specs.php}} of project Lemure) and of Twitter datasets (\textit{e.g.} the \texttt{infochimps collections}\footnote{\url{http://www.infochimps.com/collections/twitter-census}}).
But dedicated web forum benchmark datasets are scarce.
Those that exist are usually issued from a single forum website (\textit{e.g.} the \texttt{boards.ie Forums Dataset}\footnote{\url{http://www.icwsm.org/2012/submitting/datasets/}} based on \href{http://www.boards.ie}{boards.ie} website or the \texttt{Ancestry.com Forum Dataset}\footnote{\url{http://www.cs.cmu.edu/~jelsas/data/ancestry.com/}}, based on \href{http://www.ancestry.com}{ancestry.com} website).
This is due to the diverse and ever changing structure of the websites hosting the discussions and copyright problems.
Each host website has its own license on the user-produced data, which is not always clearly stipulated.
% and which hinders the sharing of the dataset.
This leads researchers to develop their own house-bred parsers and create their own datasets.
These datasets are rarely shared with the community, which poses problems when testing new proposals and comparing to existing approaches.

%\textcolor{red}{Summarize the problems:
%Software:
%- oriented social news and/or twitter;
%- does not take into account the underlying social network;
%Dataset:
%- suffer from copyright;
%- single website.}

\subsection{Introducing CommentWatcher}

We address these issues by introducing \texttt{CommentWatcher}, an open source web-based platform for analyzing discussion on web forums.
\texttt{CommentWatcher} was designed having in mind two types of users: the forum analyst, who seeks to understand the main topics of discussion and the social interactions between users, and the researcher who needs a benchmark to test his/her proposed approaches.
Using \texttt{CommentWatcher}, the researcher can create forum discussions benchmarks without worrying for copyright issues, since the platform is open source and the text itself is not distributed (each researcher can locally recreate the benchmark dataset).

%\texttt{CommentWatcher} features mass downloading of forums from supported websites, topic extraction and visualization and extracting the underlying social network.
When building \texttt{CommentWatcher} we address the challenges that arise from retrieving forums from multiple web sources.
Not only these sources are profoundly heterogeneous in structure, but they tend to change often and render parsers obsolete.
We implement a parser architecture which is independent from the website structure and allows simple on-the-fly adding of new sources and updating the existing ones.
\texttt{CommentWatcher} also supports mass fetching of forums from supported sources by using keyword search on the internet, extracting discussion topics, creating the underlying social network structure of users and visualizing it in relation with the extracted topics.
%The platform was developed using web technologies because it is intended to be put at the disposal of the community both as source code and as a web service.

During the demonstration, the participants will be able to interact with \texttt{CommentWatcher} in a normal browser window, through the tools web interface.
The tool itself will be hosted and executed on its dedicated machine, located at the ERIC laboratory.
The tools capabilities will be illustrated by showing the participants, on-live, (a) how multiple discussion forums can be fetched by searching the web using keywords, (b) apply topics extraction algorithms and tweak their parameters, (c) visualize the extracted topic as a expression cloud and their temporal evolution and (d) visualize the social network constructed starting from the initial forums.

\section{Platform Design}

In this section, we describe the software technologies used in developing \texttt{CommentWatcher}, the general architecture and the different components to highlight their aim and the way they interact.

\subsection{Software technologies}

\texttt{CommentWatcher} is written using Java Servlets for server-side computing and Java Server Page for the dymanic webpage generation.
The support for fetching forums discussions from websites is implemented using the XLS Transformation technology.
New websites can be added dynamically, without changing the source code.
A MySql database is used for storing forum structure, user characteristics and the text.
The visualization is performed client-side into a Java Applet.

\subsection{Platform architecture}

\begin{figure}[htb]
	\centering
	\includegraphics[width=0.45\textwidth]{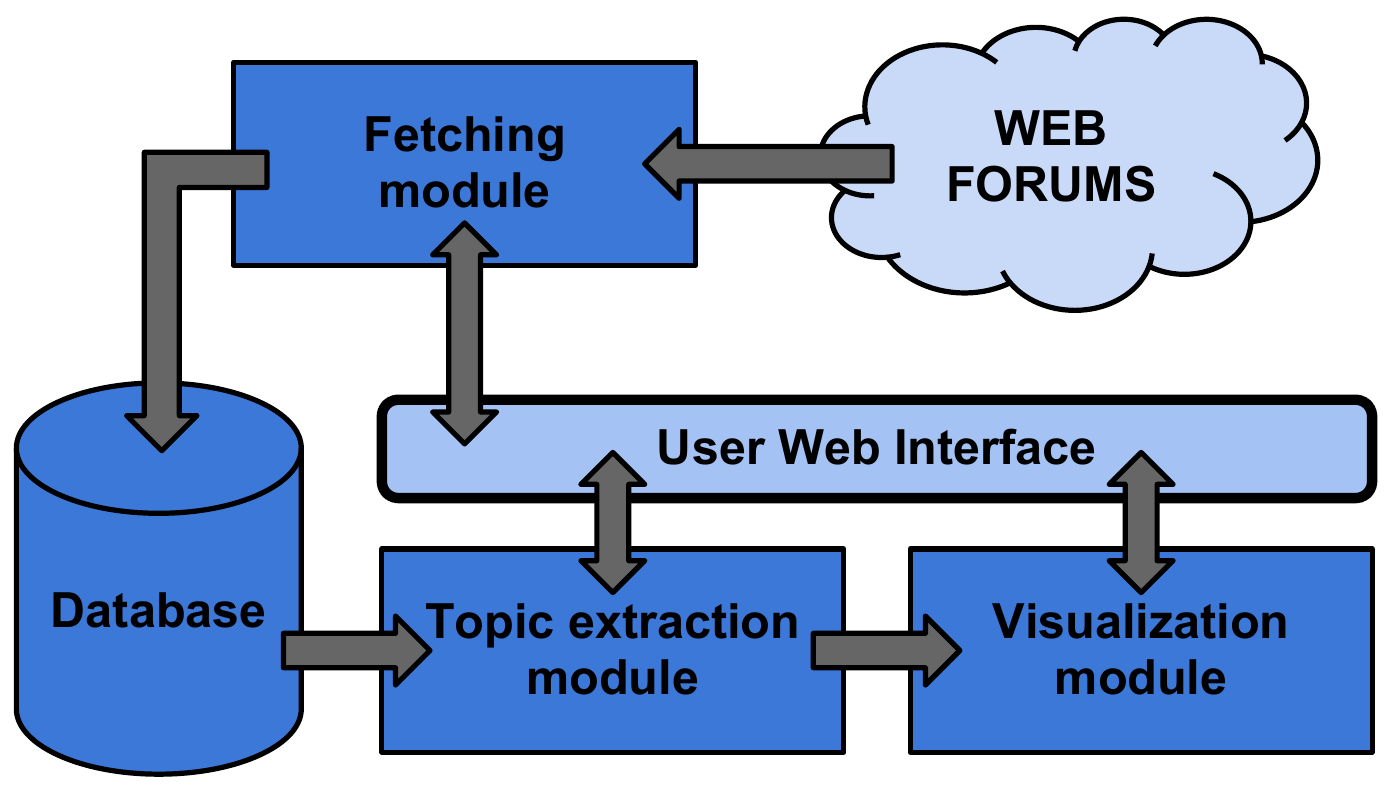}
	\caption{\texttt{CommentWatcher:} overview of the platform's architecture.}
	\label{fig:main-architecture}
\end{figure}

The application has three main modules, interconnected as shown in Figure~\ref{fig:main-architecture}.
The \textit{fetching module} deals with downloading the forums, parsing the web pages and storing the data into the database.
Optionally, it can perform a keyword web search to find forums that can be fetched.
The \textit{topic extraction module} performs topic extraction using an algorithm implemented as a library on a selection of forums.
The \textit{visualization module} has two views: (i) topic visualization as an expression cloud and as a temporal evolution graphic and (ii) social network visualization.

\subsection{The fetching module}

\begin{figure*}[!t]
  \centering
  \subfloat[] {
  	\label{sub-fig:fetching-module-schema}
  	\includegraphics[height=0.22\textheight]{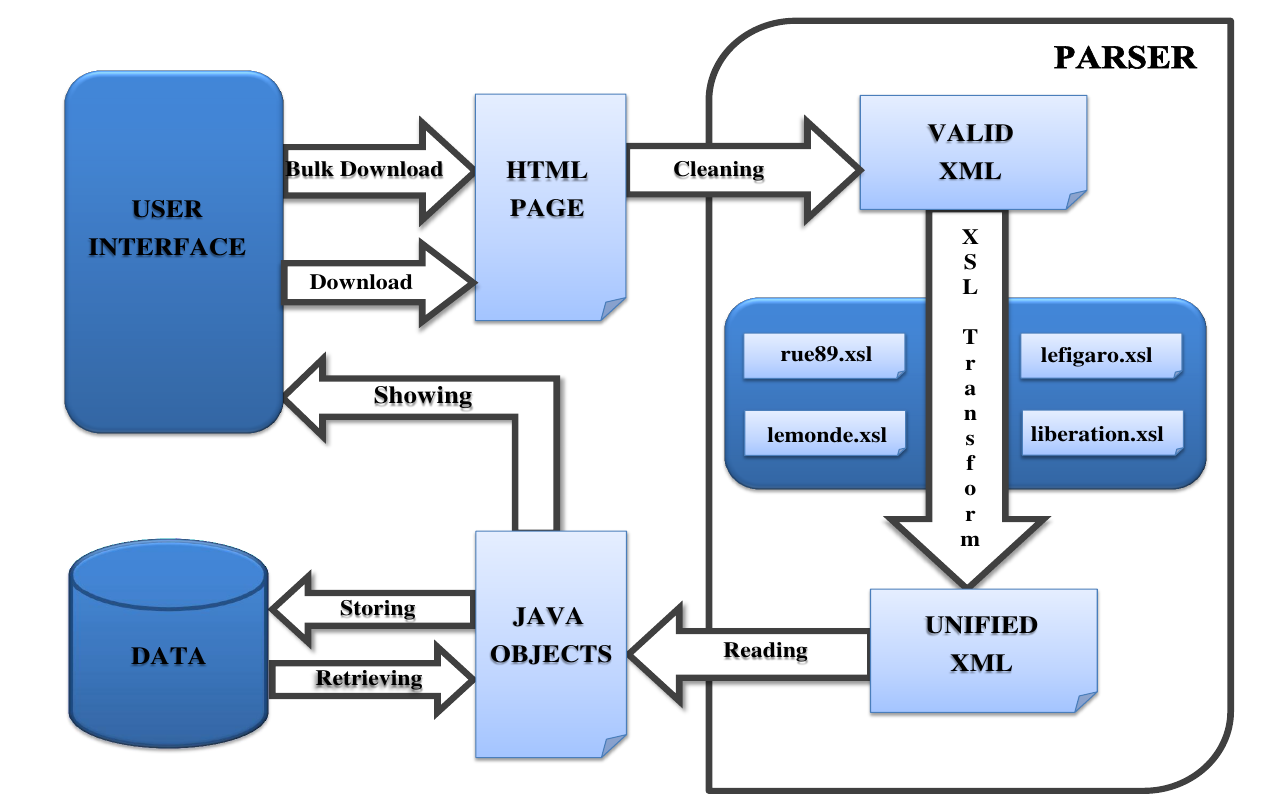}
  }         
%  \hspace{2cm}
  \hfill
  \subfloat[]{
  	\label{sub-fig:bulk-fetching-screenshot}
  	\includegraphics[height=0.22\textheight]{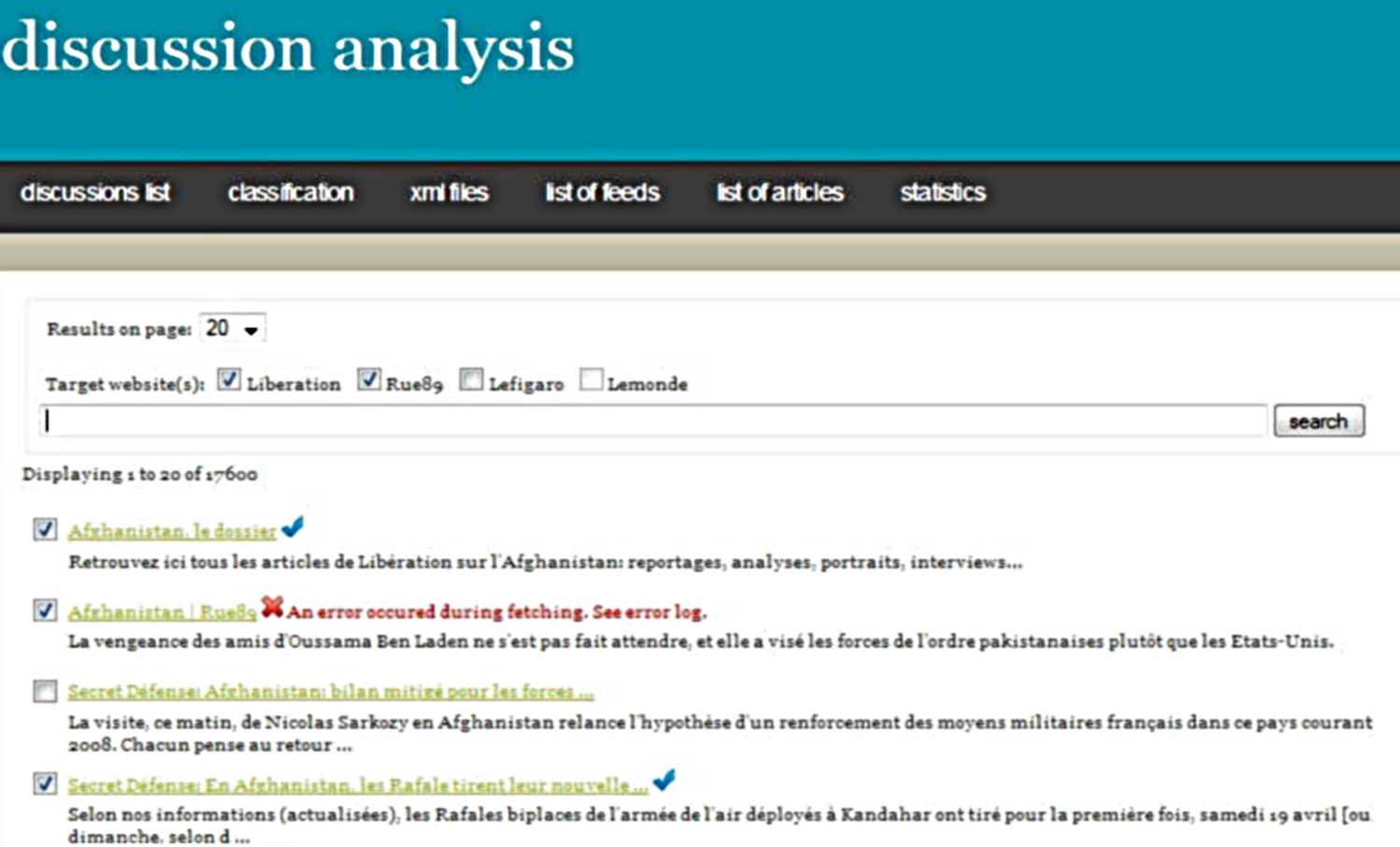}
  }
  
  \caption{The design of the fetching module (a) and a screenshot of the keyword mass fetching process (b)}
  \label{fig:chap3-algo-evolution}
\end{figure*}

This module deals with downloading, parsing and importing the forum data into the application.
The main difficulty when parsing web pages is that the structure of each page is different.
What is more, the structure of a certain web page tends to change over time.
With \texttt{CommentWatcher} we have designed and implemented a meta-parser, which is independent on the website.
The actual adaptation of the parser to a specific page is done using an external definition file, implemented in XSLT, a standardized and well documented language.
Therefore, adding support for new websites or modifying existing ones boils down to just adding or modifying definition files, without any change in the parser's source code.

The design schema of the fetching module, as well as its interactions with the user interface and the database, are given in Figure~\ref{sub-fig:fetching-module-schema}.
The download action specifies the URL of a forum to be downloaded. 
The bulk download follows the same idea, but a keyword web search is performed using the Bing API and all results from supported websites are downloaded.
A screenshot of the keyword web search and mass fetching is given in Figure~\ref{sub-fig:bulk-fetching-screenshot}.
The specified page will be downloaded in raw HTML format which will undergo cleaning, XSL transformation and deserialization. 
The process of cleaning implies transforming the HTML document into a well formed XML. 
In the following step, the XSL transformation is applied to the valid XML document using one of the XSLT definition files of the supported websites. 
The result of the transformation is an XML document, which uses the same XML schema for all supported websites. 
The required data is then deserialized into Java objects, which can be further on stored in and retrieved from the database.

The advantages of implementing such a parsing process are that it is simple, reliable, easy to understand and modify.
Furthermore, it does not hard-code the website's structure and it allows adding new supported websites on-the-fly.

\subsection{Topic extraction and textual classification}

This module allows extracting topics from texts from a selection of forums, already fetched in the database.
The design is modular, the extraction itself being performed by external libraries.
The text from selected forums is prepared and packaged in the format required by the topic extraction library and then passed to the library.
The user interface allows setting the parameters for each library.
Once the extraction is finished, the results are saved into an XML document, which has the same format for all topic extraction libraries.
The XML document contains the expressions associated to each topic and their scores.

At the present, \texttt{CommentWatcher} supports two topic extraction algorithms, provided by two libraries: Topical N-Grams~\cite{WAN07} provided by the Mallet Toolkit library\cite{MCC02} and CKP~\cite{RIZ10}, provided by the CKP library.
Topical NGrams is a graphical model algorithms, which models topics as distributions of probabilities over n-grams.
CKP uses overlapping textual clustering (one text can belong to multiple clusters) and considers each cluster of the partition as a topic.
The expressions stored in the XML result document are either (i) the resulted n-grams (for Topical NGrams) or (ii) the frequent expressions (for CKP).
Their score is (i) the probability to which an n-gram is associated to a topic (for Topical NGrams) or (ii) $1 - d(e_i, \mu)$, where $d(e_i, \mu)$ is the normalized distance between the frequent expression $e_i$ and the topic's centroid $\mu$ (for CKP).
Support for new algorithms and libraries can be added easily, but it requires writing adapters for the inputs and outputs.

%\subsection{Visualization}
\subsection{Visualization}

\begin{figure}[!t]
	\centering
	\includegraphics[height=0.223\textheight]{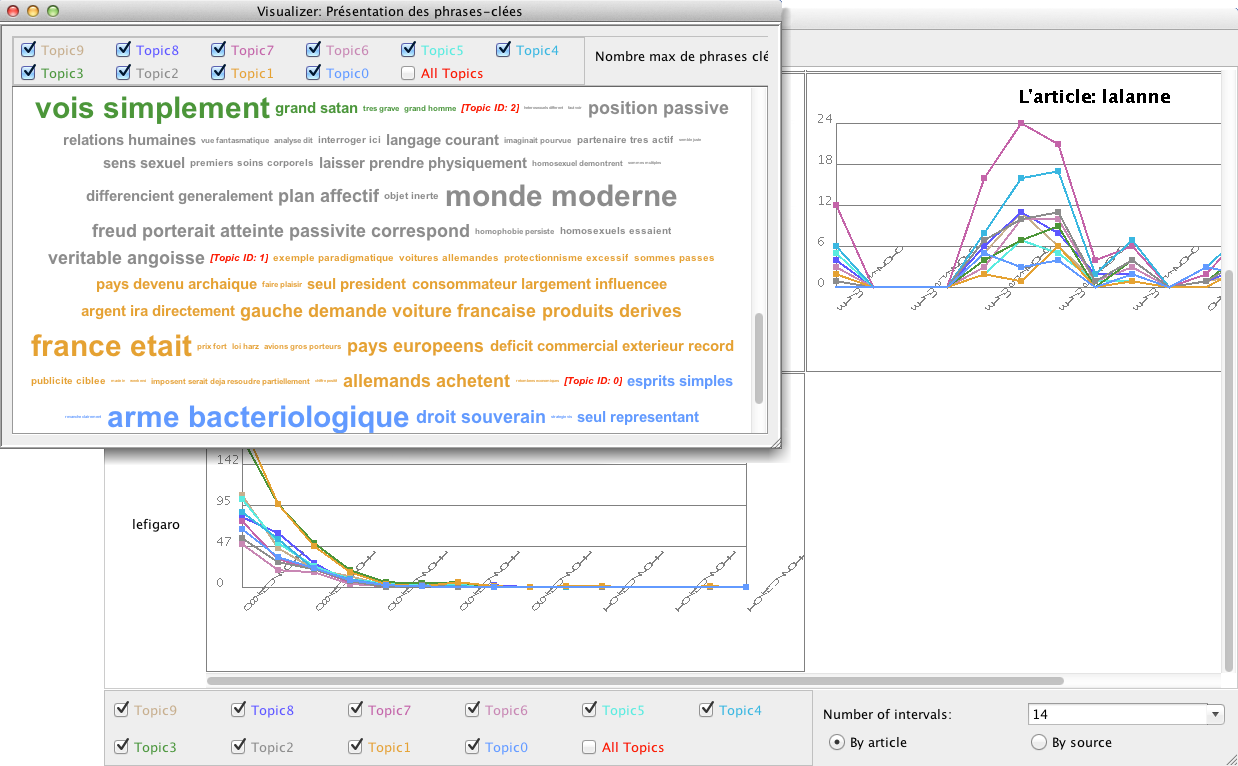}
	\caption{The expression cloud visualization of topics and their temporal evolution.}
	\label{fig:topic-visualization}
\end{figure}

The visualization module is designed to help the user to quickly understand the extracted topics and visualize their temporal evolution.
It is the only module that is executed client-side, in a Java Applet.
%The XML object resulted from the topics extraction algorithm is sent to the applet.
After the XML object resulting from the topic extraction is loaded by the applet, two visualizations are available: the expression cloud and the temporal evolution graphic.
Figure~\ref{fig:topic-visualization} shows a screenshot with the two visualizations.
The expression cloud visualization is similar to the word cloud visualization, which the exception that it uses the expressions generated at the topic extraction module and their sizes are proportional with their score.
The temporal evolution graphic portrays the popularity of each topic over the period of time.
The time is discretized in a configurable number of intervals, the user posts associated to each topic in each interval are counted and graphics are generated for each forum or for each hosting website.

\paragraph{Social network visualization}
To facilitate the exploration of the interactions between the members of the forum, we compute a visualization of the underlying social network.
The network is colored according to the topics on which the users are interacting.
We construct the social network as a labeled multidigraph, as shown in~\cite{FOR11}.
We map the network nodes on the authors of messages.
We add an arc labeled with the topic between two nodes when there is, between the two users, at least one direct reply belonging to the respective topic.
% and the arcs on the reply relations between messages.
We 
%correlate the obtained oriented graph structure with topics constructed by the topic extraction module and 
further enrich the network with user's features as the number of posts, the number of topics a user participates in, the number of threads a user participates in, \textit{etc.}
%, the number of threads initiated by a user and the number of alone post.
Further measures are calculated on the graph, such as the weighted in- and out-degree, the betweenness centrality and the closeness centrality.

\begin{figure}[!t]
	\centering
	\includegraphics[height=0.2\textheight]{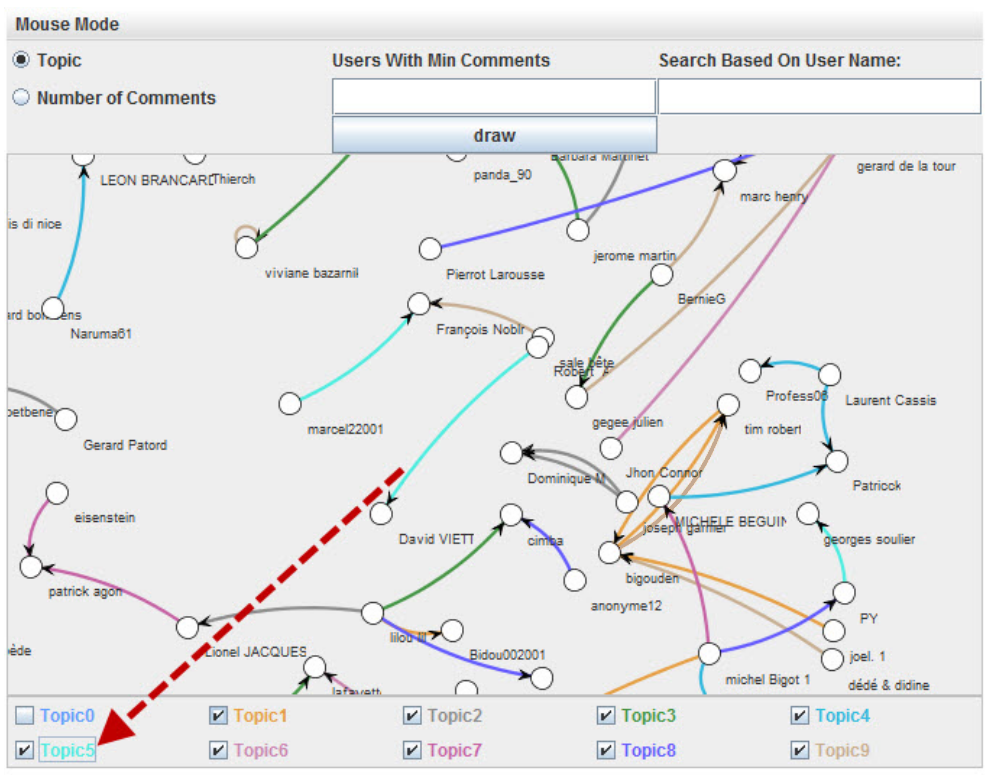}
	\caption{Visualizing the constructed social network, enriched with topical an user features. One can see, \textit{inter alia}, that the reply of ``Robert'' to ``David VIETI'' is associated to topic \#5.}
	\label{fig:social-network}
\end{figure}

Figure~\ref{fig:social-network} shows how \texttt{CommentWatcher} displays the induced social network.
The visualization is created with the Jung 
Graph 
Library\footnote{\url{http://jung.sourceforge.net}}
%The arcs are colored according to the topics and the visualization can be filtered in order to show only the network corresponding to certain topics. 
and is interactive, so nodes can be selected in order to see their features. 
Relations can also be filtered in order to show only the network corresponding to certain topics.

%\section{Demonstration}
%
%\begin{figure}[htb]
%	\centering
%	\includegraphics[height=0.2\textheight]{main-table-screenshot}
%	\caption{\texttt{CommentWatcher}'s main interface window.}
%\end{figure}
%
%\textcolor{red}{Describe what will be done during the demonstration:
%Mass-fetching of forums of articles corresponding to a search
%Fetching of one forum
%Topic extraction from the forums
%Visualizing
%Social Network}

\section{License and source code}

%\paragraph{License}
\texttt{CommentWatcher} is released under the opensource license
%GNU General Public License version 3 (
GNU GPL v3\footnote{\url{http://www.gnu.org/licenses/}}.
The individual topic extraction and textual clustering software packages are the objects of their respective licenses.
The present version of \texttt{CommentWatcher} comes with two Natural Language Processing toolkits: the Mallet Toolkit\cite{MCC02} v2.0.7, released under the open source Common Public License, and CKP\cite{RIZ10} v0.2, released under the GNU GPL v3.
%General Public License version 3.
The install files and the source code of \texttt{CommentWatcher} is available through a public Mercurial repository\footnote{\url{http://eric.univ-lyon2.fr/~commentwatcher/cgi-bin/CommentWatcher.cgi/CommentWatcher/}}.

%%\paragraph{Source code and installation}
%
%The repository contains:
%\hspace{-2cm}
%\squishlist %\begin{itemize}
%	\item A Web Application Archive (WAR file) which can be used to easily install \texttt{CommentWatcher} on an Apache Tomcat Server;
%	\item An archive which contains the structure of the writable folder needed for executing \texttt{CommentWatcher};
%	\item The source code of \texttt{CommentWatcher};
%	\item A README file which describes how to install \texttt{CommentWatcher} on a Unix/Linux machine.
%\squishend %\end{itemize}

\section{Related works}

Several tools intending to extract knowledge from on-line discussions have been proposed in the recent years.

MAQSA \cite{AME12} is a system for social analytics on news that allows its users to define their own topic of interest, in order to gather related articles, identify related topics, and extract the time-line and network of comments that show who commented which article and when.

Eddi \cite{BER10} offers visualizations such as time-lines and tag clouds of topics extracted from tweets using a simple topic detection algorithm that uses a search engine as an external knowledge base.

\textit{OpinionCrawl}\footnote{\url{http://opinioncrawl.com}} is an on-line service that crawls various web-sources -- such as blogs, news, forums and Twitter -- searching for a user-defined topic and then presents key concepts as a tag cloud, provides a visualization of the temporal dynamics of the topic and performs a sentiment analysis.

SONDY \cite{GUI13} is an open-source plateform for analyzing on-line social network data. It features a data import and pre-processing service, a topic detection and trends analysis service, as well as a service for the interactive exploration of the corresponding networks (\textit{i.e.}, active authors for the considered topic(s)).

The aforementioned tools are limited for various reasons. They are either proprietary softwares and thus can't be extended for scientific purposes or can't directly crawl web sources and can only be used to analyze formatted datasets provided by the user. \texttt{CommentWatcher} intends to provide researchers with an open-source extendable tool that permits to crawl the web and build datasets that suit their needs. 

%\textcolor{DarkViolet}{
%MAQSA: A System for Social Analytics on News~\cite{AME12} is oriented towards news articles and tweets.
%en gros y a une extraction de thématique dans les news
%+ extration d'entité
%après ca monte un réseau de thématique si je me souviens bien
%et puis ca géolocalise les gens qui publient des commentaires
%enfin, ca cartographie*}

%\textcolor{red}{Take more references from the related works of the previous article.}
%
%\textcolor{red}{Describe a little plateforms that deal with social networks without treating forum data.}

\section{Conclusion and future works}

In this paper we have presented \texttt{CommentWatcher}, an open source web-based platform for analyzing discussions on web forums.
Our tool is designed for both end-users, as well as for researchers.
End-users have at their disposal an easy to use, integrated tool that allows retrieving forum discussion from multiple websites, performs topic extraction to identify the main discussion topics and provides an expression cloud visualization to identify the most important expressions associated to each topic.
The temporal popularity of topics can be evaluated using an evolution graphic.
\texttt{CommentWatcher} also features extracting the underlying social network by using the direct citation links between users.
The visualization of the social network is interactive, features of nodes can be visualized and relations can be filtered to show only the network corresponding to a certain topic.
For researchers, \texttt{CommentWatcher} tackles the problem of creating multi source web forum datasets, thanks to its versatile parser which is independent of the structure of webpages.
Support for new websites can be added on-the-fly.
It can also solve the problem of copyright when sharing forum datasets, since no text is distributed and each researcher can easily recreate the dataset.
As future work, we intend to add a credential mechanism and transform \texttt{CommentWatcher} into a multiuser tool.
We consider implementing topic evaluation based on ontologies of concepts and a better plotting of the social network by using force-directed graph drawing.
%can be coupled with the extracted topics so that only arcs
%that runs on a web server and interacts through a browser
%Conclusions, use cases.
%It will be available for the use of the community.
%
%Perspectives
%Need to add a credential mechanism.
%The work of DMKM and M1 students.

\bibliographystyle{abbrv}
\bibliography{sigproc}  % sigproc.bib is the name of the 

\balancecolumns
% That's all folks!
\end{document}